\documentclass[sigconf]{acmart}
\setcopyright{none}
\renewcommand\footnotetextcopyrightpermission[1]{}
\usepackage{enumitem}
\usepackage{tikz}
\usepackage{pgf-pie}
\usepackage{subcaption}

\usepackage{algorithm}
\usepackage{algpseudocode}

\usepackage{amsmath}
\usepackage{mathtools}
\usepackage{amsthm}
\usepackage{fixmath}
\usepackage{amsmath}
\usepackage{amsfonts}
\usepackage{pifont} 
\usepackage{xcolor} 
\usepackage{mathtools}

\usepackage{multicol}
\usepackage{multirow}

\AtBeginDocument{%
  \providecommand\BibTeX{{%
    \normalfont B\kern-0.5em{\scshape i\kern-0.25em b}\kern-0.8em\TeX}}}
    
\copyrightyear{2024}
\acmYear{2024}
\setcopyright{rightsretained}
\acmConference[ISLPED '24]{Proceedings of the ACM/IEEE International Symposium on Low Power Electronics and Design}{August 5--7, 2024}{Newport Beach, CA, USA}
\acmBooktitle{Proceedings of the ACM/IEEE International Symposium on Low Power Electronics and Design (ISLPED '24), August 5--7, 2024, Newport Beach, CA, USA}
\acmDOI{10.1145/3665314.3670843}
\acmISBN{979-8-4007-0688-2/24/08}

\begin{document}
\pagestyle{plain}

\title{PEANO-ViT: Power-Efficient Approximations of Non-Linearities in Vision Transformers}



\author{Mohammad Erfan Sadeghi}

\affiliation{%
  \institution{University of Southern California}
  \city{Los Angeles}
  \state{California}
  \country{USA}
  \postcode{90089}
}
\email{sadeghim@usc.edu}

\author{Arash Fayyazi}
\affiliation{%
  \institution{University of Southern California}
  \city{Los Angeles}
  \state{California}
  \country{USA}
  \postcode{90089}
}
\email{fayyazi@usc.edu}

\author{Seyedarmin Azizi}
\affiliation{%
  \institution{University of Southern California}
  \city{Los Angeles}
  \state{California}
  \country{USA}
  \postcode{90089}
}
\email{seyedarm@usc.edu}

\author{Massoud Pedram}
\affiliation{%
  \institution{University of Southern California}
  \city{Los Angeles}
  \state{California}
  \country{USA}
  \postcode{90089}
}
\email{pedram@usc.edu}








\begin{abstract}
The deployment of Vision Transformers (ViTs) on hardware platforms, specially Field-Programmable Gate Arrays (FPGAs), presents many challenges, which are mainly due to the substantial computational and power requirements of their non-linear functions, notably layer normalization, softmax, and Gaussian Error Linear Unit (GELU). These critical functions pose significant obstacles to efficient hardware implementation due to their complex mathematical operations and the inherent resource count and architectural limitations of FPGAs. PEANO-ViT offers a novel approach to streamlining the implementation of the layer normalization layer by introducing a division-free technique that simultaneously approximates the division and square root function. Additionally, PEANO-ViT provides a multi-scale division strategy to eliminate division operations in the softmax layer, aided by a Padé-based approximation for the exponential function. Finally, PEANO-ViT introduces a piece-wise linear approximation for the GELU function, carefully designed to bypass the computationally intensive operations associated with GELU. In our comprehensive evaluations, PEANO-ViT exhibits minimal accuracy degradation (\(\leq 0.5\%\) for DeiT-B) while significantly enhancing power efficiency, achieving improvements of \(1.91\times\), \(1.39\times\), and \(8.01\times\) for layer normalization, softmax, and GELU, respectively. This improvement is achieved through substantial reductions in DSP, LUT, and register counts for these non-linear operations. Consequently, PEANO-ViT enables efficient deployment of Vision Transformers on resource- and power-constrained FPGA platforms.

\end{abstract}

\maketitle

\section{Introduction}
The landscape of computer vision has been fundamentally transformed with the advent of deep learning, among which Vision Transformers (ViTs) \cite{DBLP:conf/iclr/DosovitskiyB0WZ21, DBLP:conf/icml/TouvronCDMSJ21, DBLP:conf/iccv/LiuL00W0LG21} have emerged as a particularly promising approach. Unlike traditional convolutional neural networks (CNNs) that rely on local receptive fields, ViTs leverage the power of self-attention mechanisms to capture global dependencies within an image, enabling a more comprehensive understanding of visual data. This capability has placed ViTs at the forefront of research, demonstrating state-of-the-art performance across a wide range of tasks in computer vision. Overall, deep learning has revolutionized various domains by providing robust algorithms capable of learning complex patterns from large datasets, thus enabling unprecedented advancements in the application of artificial intelligence across numerous fields, from healthcare \cite{razmara2024feverdetectioninfraredthermography} to recommendation systems to scientific research.

ViTs rely on a series of identical encoder blocks to progressively extract complex features from an image. These encoder blocks consist of two principal components: Multi-headed Attention (MHA) and Feed-Forward Network (FFN), each prefaced with a layer normalization block. Embedded within MHA and FFN are linear layers, GELU, and softmax, integrated via two residual connections that bookend the normalization stages. The output of the final encoder block goes through a classifier to obtain the class predictions.


Despite their exceptional performance, ViTs face significant challenges for practical deployment due to their extensive parameter count and considerable computational demands. A wide range of methods has been explored to improve the efficiency of ViTs, including approaches like quantization \cite{DBLP:conf/nips/LiuWHZMG21}, pruning \cite{DBLP:conf/aaai/0004HWCCC22}, and low-rank approximations \cite{azizi2024memoryefficient}. However, the deployment of ViTs in practical applications, especially on hardware platforms such as Field-Programmable Gate Arrays (FPGAs), presents fundamental challenges. Among these, the non-linear layers—layer normalization, softmax, and GELU—integral to the architecture of ViTs, stand out. While crucial for the network's ability to model complex patterns, these functions are computationally intensive and thus present a critical challenge for the efficient implementation on FPGAs. 


Our research delivers two key contributions. Firstly, we introduce PEANO-ViT, a novel approach that utilizes hardware-optimized approximation techniques for the non-linear functions within ViTs. 
Our approach in PEANO-ViT offers a comprehensive solution to the challenges posed by implementing key functions in ViTs on FPGA platforms. By leveraging innovative techniques such as the Padé-based approximation for the exponential function and incorporating bit manipulation operations for efficient division in the softmax layer, we strive for a well-balanced and resource-efficient implementation that prioritizes performance and resource conservation. The layer normalization implementation effectively tackles computational challenges by approximating the reciprocal of the square root, $\frac{1}{\sqrt{x}}$, in a novel manner. Furthermore, our adoption of a piece-wise linear approximation for GELU not only minimizes resource usage but also closely preserves the original function's behavior. 
Secondly, we demonstrate through comprehensive experiments that PEANO-ViT enables the efficient execution of ViTs on FPGAs, with minimal impact on accuracy and significant improvements in computational efficiency and power consumption.
\section{Related Work}
\label{sec:rel_work}
Transformers \cite{DBLP:conf/nips/VaswaniSPUJGKP17}, originally developed for tackling long sequences in natural language processing tasks, served as the inspiration behind ViT \cite{DBLP:conf/iclr/DosovitskiyB0WZ21} for computer vision applications. ViTs achieve impressive results by processing images as sequences of tokens and leveraging the power of self-attention. However, while crucial for performance, the core non-linear functions in ViTs – softmax, GELU, and layer normalization – are computationally expensive and hinder efficient hardware implementation. Several studies have explored hardware-efficient strategies for these layers, presenting various approximation techniques that balance approximation accuracy with computational cost. Their characteristic in comparison to PEANO-ViT is summarized in Table \ref{tab:comparison}. The calculations for basic layer normalization, softmax, and GELU are depicted in equations (\ref{eq:LN}-\ref{eq:approx_gelu}), respectively. In equation \ref{eq:LN}, \(\gamma\) and \(\beta\) are learnable parameters while \(\mu\) and \(\sigma\) represents the average and variance of input data of the layer normalization function.
\begin{equation}
\text{LayerNorm}(x_i) = \frac{x_i-\mu}{\sigma} * \gamma + \beta
\label{eq:LN}
\end{equation}

\begin{equation}
\text{Softmax}(x_i) = \frac{e^{x_i}}{\sum{e^{x_i}}}
\label{eq:SFTX}
\end{equation}

\begin{equation}
\text{GELU}(x) \approx 0.5 x \left(1 + \tanh\left[\sqrt{\frac{2}{\pi}} \left(x + 0.044715 x^3\right)\right]\right)
\label{eq:approx_gelu}
\end{equation}

\definecolor{darkergreen}{RGB}{0,150,0}

\begin{table*}[tb]
\centering
\scriptsize
\caption{Comparison of state-of-the-art methods for implementation of non-linear layers}
\vspace{-4mm}
\label{tab:comparison}
\resizebox{0.80\textwidth}{!}{
\begin{tabular}{c c c c c c}

\toprule
\multirow{2}{*}{\textbf{Approach}}   & \textbf{Layer normalization}   & \textbf{Softmax} & \textbf{GELU} &\textbf{All division-free}  &\textbf{Accuracy and resource aware}       \\
{}                                  & \textbf{approximation}              & \textbf{approximation}  & \textbf{approximation} & \textbf{ approximations} & \textbf{flexible approximations}\\

\midrule[\heavyrulewidth]
\textbf{Softermax \cite{DBLP:conf/dac/StevensVDKR21}}  & \textcolor{red}{\ding{55}}  & 
\textcolor{darkergreen}{\ding{51}}  & \textcolor{red}{\ding{55}}  & \textcolor{red}{\ding{55}} & \textcolor{red}{\ding{55}}    \\

\midrule[\heavyrulewidth]
\textbf{Koca et al.\cite{DBLP:conf/iscas/KocaDC23}}  & \textcolor{red}{\ding{55}}  & \textcolor{darkergreen}{\ding{51}}   & \textcolor{red}{\ding{55}} & \textcolor{darkergreen}{\ding{51}} & \textcolor{red}{\ding{55}}    \\

\midrule[\heavyrulewidth]
\textbf{Peltekis et al.\cite{DBLP:journals/corr/abs-2402-10118}}  & \textcolor{red}{\ding{55}}  & 
\textcolor{darkergreen}{\ding{51}}   & \textcolor{darkergreen}{\ding{51}} & \textcolor{darkergreen}{\ding{51}} & \textcolor{red}{\ding{55}}    \\

\midrule[\heavyrulewidth]
\textbf{SOLE \cite{DBLP:conf/iccad/WangZSSL23}}  & \textcolor{darkergreen}{\ding{51}}  & 
\textcolor{darkergreen}{\ding{51}}   & \textcolor{red}{\ding{55}} &\textcolor{red}{\ding{55}} & \textcolor{darkergreen}{\ding{51}}  \\

\midrule[\heavyrulewidth]
\textbf{Li et al.\cite{li2023high}}  & \textcolor{red}{\ding{55}}  & 
\textcolor{darkergreen}{\ding{51}}   & \textcolor{darkergreen}{\ding{51}} & \textcolor{darkergreen}{\ding{51}} & \textcolor{red}{\ding{55}}    \\

\midrule[\heavyrulewidth]
\textbf{LTrans-OPU \cite{DBLP:conf/fpl/BaiZZZCYW23}}  & \textcolor{darkergreen}{\ding{51}}  & \textcolor{darkergreen}{\ding{51}}
&\textcolor{darkergreen}{\ding{51}}   & \textcolor{darkergreen}{\ding{51}}  &  \textcolor{red}{\ding{55}}   \\

\midrule[\heavyrulewidth]
\textbf{PEANO-ViT (Ours)}  & \textcolor{darkergreen}{\ding{51}}  & 
\textcolor{darkergreen}{\ding{51}}   & \textcolor{darkergreen}{\ding{51}} & \textcolor{darkergreen}{\ding{51}} & \textcolor{darkergreen}{\ding{51}}   \\

\bottomrule
\end{tabular}}
\vspace{-3mm}
\end{table*}

\subsection{Softmax Implementations}
\label{subsec:rw_softmax}
The implementation of the softmax layer has emerged as a focal point of research, with numerous studies dedicated to optimizing its efficiency through various approximation techniques. The main challenges for an efficient implementation of softmax on hardware platforms arise from the non-linear function of \(e^{x}\) and the final division operation for normalizing the output values. Previous research efforts, such as those by \cite{DBLP:conf/dac/StevensVDKR21} targeted the efficient calculation of exponential function and but were hindered by the costly division operation. In contrast, studies by \cite{DBLP:conf/iscas/KocaDC23}, \cite{DBLP:conf/iccad/WangZSSL23}, and \cite{li2023high} adopted bit manipulation techniques to simplify the exponential function approximation and eliminate the need for division. Although these methods are beneficial for reducing computational demands and are well-suited for hardware implementation, they still have a high computational complexity due to their inherently iterative nature, causing increased inference latency. 

   

   

\subsection{Layer Normalization Implementations} 
\label{subsec:rw_LN}

For hardware implementation of layer normalization, significant hurdles include the efficient approximation of the square root function and managing division operations. 
the approach introduced in \cite{DBLP:conf/iccad/WangZSSL23} tackles the division operation issue but continues to employ the exact yet resource-intensive formula of square root, resulting in lower throughput. 

\subsection{GELU Implementations} 
\label{subsec:rw_Gelu}

Beyond layer normalization and softmax, the GELU function's approximation also poses a significant challenge in the hardware deployment of ViTs. This is due to its intricate non-linear nature, which necessitates the execution of the \(tanh(x)\) function alongside polynomial calculations. Authors of \cite{li2023high} have explored the approximation of the GELU function by simplifying the non-linear \(2^{x}\) function using bit manipulation operations. Additionally, \cite{DBLP:journals/corr/abs-2402-10118} has presented an innovative method that leverages existing softmax hardware to facilitate GELU computations. While these approaches are designed to be hardware-efficient and minimize resource consumption, the computational latency remains a concern. This is due to the iterative nature of some of the bit manipulation operations in \cite{li2023high}, and the use of non-optimized hardware for GELU in \cite{DBLP:journals/corr/abs-2402-10118}. 

%

\section{Methodology}
\label{sec:method}
In this section, we describe the techniques utilized to approximate the layer normalization, softmax, and GELU functions. Our emphasis was on developing methods that avoid divisions and ensure compatibility with hardware implementations while also aiming to preserve the accuracy of the model as much as possible.

\subsection{Layer Normalization}
\label{sec:method_ln}
As described in subsection \ref{subsec:rw_LN}, the main challenges of implementing layer normalization on hardware platforms such as FPGAs are the non-linear square root function and the costly division operation. Inspired by SOLE \cite{DBLP:conf/iccad/WangZSSL23}, we propose a method to directly approximate \( \frac{1}{\sqrt{X}} \). We start with the following identities:
\begin{equation}
    \frac{1}{\sqrt{X}} = 2^{\log_{2}{\frac{1}{\sqrt{X}}}},  \quad  \log_{2}{\frac{1}{\sqrt{X}}} = \frac{-1}{2} \log_{2}{X}
\end{equation}

Based on \cite{DBLP:conf/iccad/WangZSSL23}, we use equations (\ref{eq:SOLE1}-\ref{eq:SOLE2}) to approximate \(\log_{2}{X}\), in which \(k_x\) is the leading '1' bit of \(X\) and $x \in [0, 1)$: 
\begin{equation}\label{eq:SOLE1}
    X  = \sum_{i=0}^{n-1} 2^i b_i = 2^{k_x} + \sum_{i=0}^{k_x-1} 2^i b_i = 2^{k_x} (1+x)
\end{equation}
\begin{equation}\label{eq:SOLE2}
    \log_{2}{X}  \approx k_x + x
\end{equation}
Therefore, we can have the following approximation:
\begin{equation}\label{eq:SOLE4}
     \frac{1}{\sqrt{X}} \approx 2^{\frac{-(k_x + x)}{2}}
\end{equation}
Calculating the \(2^{\frac{-(k_x + x)}{2}}\) term is the only step remaining. We note that \(2^{\alpha} = 2^{u} * 2^{v}\) in which \(u\) is an integer number and $v \in [0, 1)$. To avoid calculating \(2^{v}\), we keep the top \(m\) bits of \(v\)'s binary representation as \(v \approx \tilde{v} = (0.v_{-1}\dots v_{-m})_2\) and pre-store \(2^{(0.0\ldots0)_2}\) up to \(2^{(0.1\ldots1)_2}\). Since \({u}\) is an integer number, \(2^{u}\) can be implemented using the shift operation. Thus, the approximation of \(\frac{1}{\sqrt{X}}\) can be obtained from  two equations below:
\begin{equation}\label{eq:SOLE5}
     2^{\frac{-(k_x + x)}{2}} = 2^{u} \cdot 2^{v}
\end{equation}
\begin{equation}\label{eq:SOLE6}
     \frac{1}{\sqrt{X}} \approx 2^{\tilde{v}} << {u}
\end{equation}
Figure \ref{fig:rsqrt2} shows the \(\frac{1}{\sqrt{X}}\) compared to our approximation and the overall layer normalization method is described in algorithm \ref{alg:PLN}. Using these approximations, we have simultaneously tackled the two problems of efficient implementation of the square root function and approximating the division operation. It is important to highlight that \(m\), an adjustable integer parameter, enables a trade-off between the precision of the approximation and the on-chip memory requirements for storing \(2^{\tilde{v}}\). Increasing \(m\) improves the approximation accuracy at the cost of demanding more on-chip memory. This flexibility will be discussed in detail in Section \ref{subsec:flexibility}.

\begin{algorithm}[tb]
    \caption{PEANO Layer Normalization}
    \label{alg:PLN}
    \begin{algorithmic}[1]
   
        \Require $x_1, \ldots, x_n$, $\gamma$, $\beta$, $fracPow2[2^{m}] = \{2^{(0.0\ldots0)_2},\ldots,2^{(0.1\ldots1)_2}\}$
        \Ensure $y_1,\ldots, y_n$
        \State $Avg = \frac{1}{n} \sum_{i=1}^{n} x_i$ \Comment{average of inputs}
        \State $AvgSQ = \frac{1}{n} \sum_{i=1}^{n} x_i^2$ 
        \Comment{average of inputs squared}
        \State $Var = AvgSQ - Avg^2$ 
        \Comment{variance of inputs}
        \State $k_{Var} = LeadingOne(Var)$ 
        \Comment{leading '1' bit of variance}
        \State $x_{Var} = Var[k_{Var}-1:0]$ 
        \Comment{contains the bits after $k_{Var}$}
        \State $log_{2}Approx = {-(k_{Var}+x_{Var})}>>1$ 
        \State $u = \lfloor log_{2}Approx \rfloor$ 
        \State $v = u - log_{2}Approx$ 
        \State $\tilde{v} = fracBits(v,m) $ 
        \Comment{$\tilde{v}$ keeps the top m fractional bits in $v$}
        \State $recipSqrt = fracPow2[\tilde{v}] << u $ 
        \Comment{approximation of $\frac{1}{\sqrt{Var}}$}
            \For{$i=1$ {\bfseries to} $n$}
                \State $y_i = (x_i - Avg) * recipSqrt * \gamma + \beta  $
            \EndFor
        \State \Return $y_1, y_2, \ldots, y_n$
    \end{algorithmic}
\end{algorithm}

\subsection{Softmax}
\label{sub:method_softmax}
\begin{figure}[tbp]
     \centering
     \vspace{-2mm}
     \begin{subfigure}[b]{0.47\columnwidth}
         \centering
         \includegraphics[width=\textwidth]{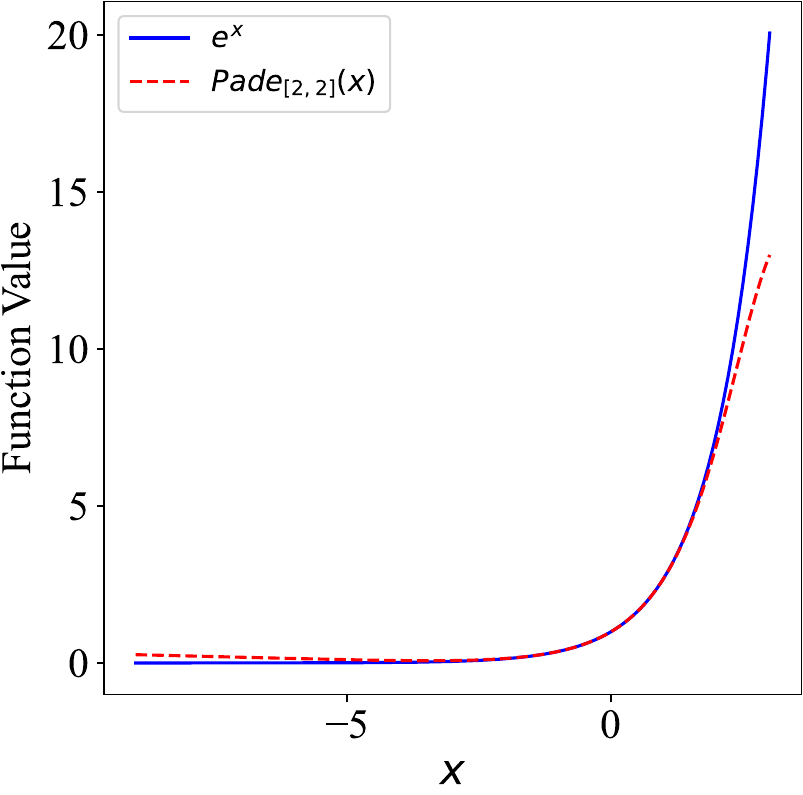}
         \caption{Exponential func.}
         \label{fig:exp}
     \end{subfigure}
     \hfill
     \begin{subfigure}[b]{0.49\columnwidth}
         \centering
         \includegraphics[width=\textwidth]{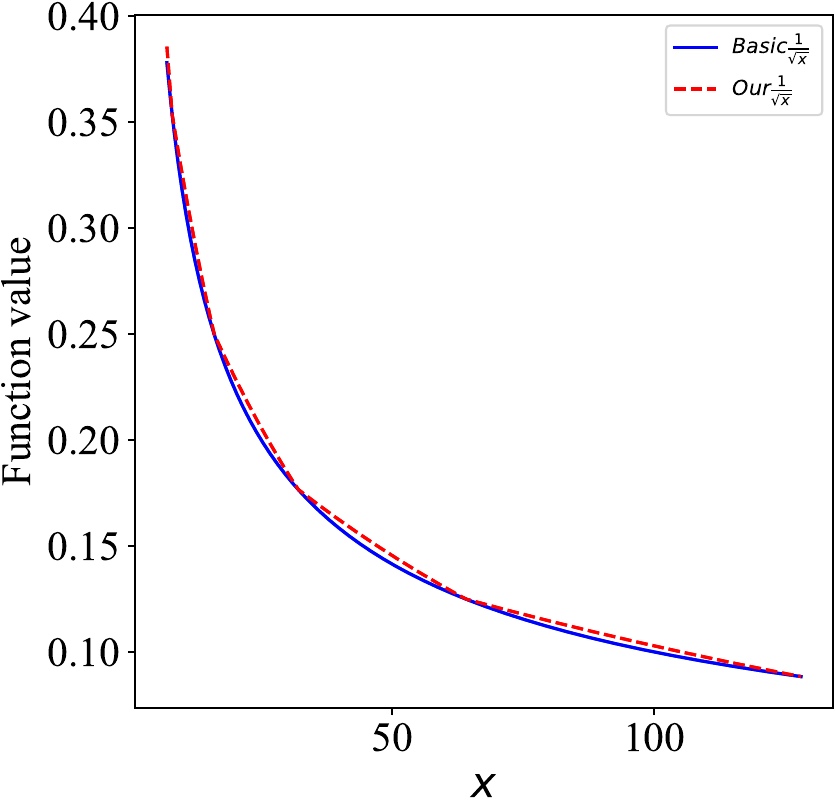}
         \caption{Reciprocal square root func.}
         \label{fig:rsqrt2}
     \end{subfigure}
     \hfill
     \begin{subfigure}[b]{0.49\columnwidth}
         \centering
         \includegraphics[width=\textwidth]{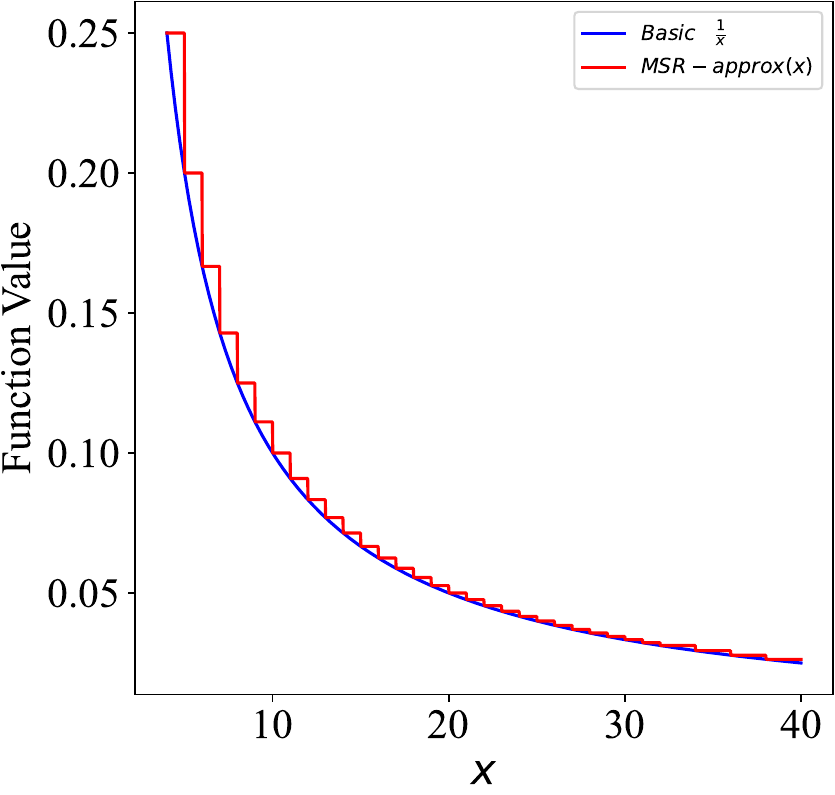}
         \caption{Reciprocal func.}
         \label{fig:rf}
     \end{subfigure}
     \hfill
     \begin{subfigure}[b]{0.46\columnwidth}
         \centering
         \includegraphics[width=\textwidth]{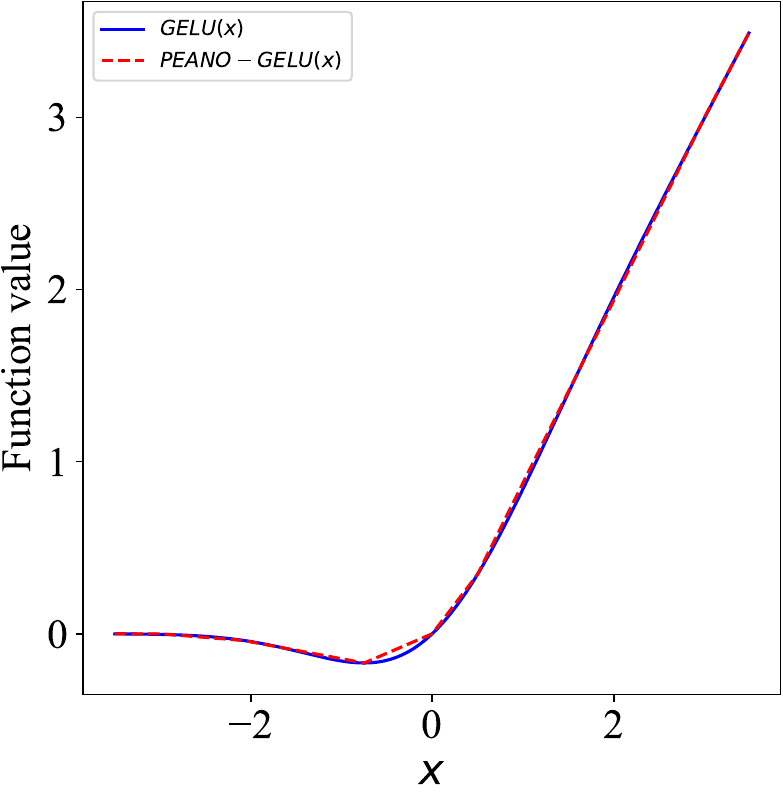}
         \caption{GELU func.}
         \label{fig:gelu}
     \end{subfigure}
     \vspace{-3mm}
        \caption{\small Comparison of standard functions with our approximations.}
        \label{fig:pade-plot}
        \vspace{-5mm}
\end{figure}

Our method for softmax approximation includes two steps. First, we introduce a Padé-based approximation for the exponential function. In the second step, we eliminate the division operations by proposing a multi-scale reciprocal approximation (MSR-approx)  method.
The Padé approximation \(Pade_{[m, n]}(x) = \frac{a_0 + a_1x + \ldots + a_{m-1}x^{m}}{b_0 + b_1x + \ldots + b_{n-1}x^{n}}\) of a function \(f(x)\) is the ratio of 2 polynomial functions. It represents a better approximation of an arbitrary nonlinear function compared to pure polynomial approximations of the same degree. For approximating the \(e^{x}\) term, we have set \( m = n = 2\) to get a Padé approximation as follows:
\begin{equation}\label{eq:Pade}
     e^{x} \approx \frac{12 + 6x + x^2}{12 - 6x + x^2}
\end{equation}
To compute the \(Pade_{[2, 2]}(x)\) approximation of \(e^{x}\), we only need to compute \(x^{2} = x \cdot x\) and \(6x = x << 2 + x << 1\) thanks to the numerator and denominator having similar functional forms. Figure \ref{fig:exp} illustrates the Pade-based approximation of the function compared to \(e^{x}\). As can be seen, the proposed approximation is very accurate for \(x \in [-3, 2]\). This observation motivated us first to add 2 to all inputs (after subtracting the maximum value) and then set \(e^{x}\) to \(0\) for the values of less than \(-3\) after the first step's calculations. Our final approximation of the exponential function is thus as follows:
\begin{equation}\label{eq:Pade softmax}
  PEANOexp(\tilde{x}) = 
  \begin{cases} 
   0 & \text{if } \tilde{x} < -3 \\
   \frac{12 + 6\tilde{x} + \tilde{x}^2}{12 - 6\tilde{x} + \tilde{x}^2} & \text{if } \tilde{x} \geq -3 
  \end{cases}
\end{equation}
Where $\tilde{x} = x - max(x_i) + 2$. The above approximation adds another division operation to the main calculation of softmax. The first division is for the computation of 
\(PEANOexp(x)\) while the second division is needed for the softmax's output normalization. Since \(\tilde{x} \in [-3, 2]\), values of the \(PEANOexp(x)\)'s denominator lie in the interval of \([4, 39]\). This motivated us to pre-store some of \(\frac{1}{x}\) values and subsequently use them to approximate the reciprocal function. However, unlike the denominator of \(PEANOexp(x)\), the denominator of the second division has a huge range of values. Therefore, pre-storing values to approximate the second division is not feasible (unless a very large lookup table is used, which would result in high memory usage.) 

To solve the aforesaid problem, we propose a multi-scale reciprocal approximation (MSR-approx) scheme for both division operations in the softmax. First we replace \(X\) (the denominator) with \(\tilde{X}\) using the equation below:
\begin{equation}
  \Tilde{X} = Scale \cdot\lfloor \frac{X}{Scale} \rfloor 
\end{equation}
And the reciprocal function approximation is described as,
\begin{equation}\label{eq: base reciprocal}
  \frac{1}{X} \approx \frac{1}{\Tilde{X}} = \frac{1}{Scale} \cdot\frac{1}{\lfloor \frac{X}{Scale} \rfloor } 
\end{equation}
Next, we force \(Scale = 2^{\alpha}\) to be an integer power of 2 so that \(\frac{1}{Scale}\) can be implemented by using a right shift by $\alpha$. This constraint also helps with the calculation of \(\lfloor \frac{X}{Scale} \rfloor\) since it simply means dropping out the $\alpha$ right bits of \(X\). The only thing we need to do is to pre-store \(\lfloor \frac{X}{Scale} \rfloor\) values, which is still problematic due to the fact that the range of \(X\) can be extremely wide for the second division operation. This arises from the assumption of fixed $\alpha$ for all X values while using a dynamic value of $\alpha$ will solve the problem of \(X\)'s large variable range as described in algorithm \ref{alg:base recip}.

Algorithm \ref{alg:base recip} shows the multi-scale approximation of the reciprocal function, which uses an adjustable integer threshold $\alpha^*$ and pre-stored values of \(\{\frac{1}{1},\ldots,\frac{1}{2^{\alpha^*+1}-1}\}\). The MSR-approx maps all values of \(X\) into the interval of \([1,2^{\alpha^*+1}-1]\) via defining a flexible Scale value, which solves the problem of the dynamic range of \(X\). For instance, if \(\alpha^*=4\) then for \(X \in [1,31]\) then \(\lfloor\frac{X}{Scale}\rfloor \in \{1,\dots,31\}\), and \(\lfloor\frac{X}{Scale}\rfloor \in \{16,\dots,31\}\) for the other values of \(X\). Hence, we only need to pre-store \(\{\frac{1}{1},\ldots,\frac{1}{31}\}\). Figure \ref{fig:rf} illustrates our MSR-approx method compared to original reciprocal function for \(\alpha^*=4\). Choosing \(\alpha^*\) is a trade-off between the accuracy of MSR-approx and the memory required for pre-storing values (see Section \ref{subsec:flexibility}). Larger \(\alpha^*\) proposes a more accurate approximation of reciprocal function while requiring larger memory for pre-stored values. The softmax using the MSR-approx scheme is presented in algorithm \ref{alg:PN-SFTMX}.

An alternative approach for improving the accuracy of the multi-scale division is to use linear interpolation between pre-stored points (instead of directly using any of these points.) For instance, if \(X = 59\) and \(\alpha^* = 4\), the scale is equal to 2, so in the basic MSR-approx method, we approximate \(\frac{1}{59}\) using \(\frac{1}{\lfloor\frac{59}{2}\rfloor} =\frac{1}{29}\). Instead, we can do linear interpolation between \(\frac{1}{29}\) and \(\frac{1}{30}\) to have a more accurate approximation of \(\frac{1}{59}\). The MSR approximation enhanced with linear interpolation (called LMSR-approx) attains superior accuracy at the expense of a slight increased resource consumption and computational cycles, illustrating a clear trade-off between accuracy and resource efficiency.

\subsection{GELU}
\label{sub:method_gelu}
PEANO-ViT uses a piece-wise linear approach to approximate the Gaussian Error Linear Unit (GELU). Unlike ViT's other non-linear functions, such as the square root and exponential functions, GELU exhibits a predominantly linear behavior across both the lower and upper extremes of its domain. Additionally, the GELU activation function maintains a narrow range of values within its non-linear region. These characteristics motivate the adoption of a piece-wise linear approximation as a highly suitable method for replicating the functionality of the GELU function.

Our method employs six breakpoints for GELU computations, resulting in seven linear segments. The initial breakpoints are set at \(x = -3\) and \(x = 3\), chosen to emulate the GELU's linear behavior as \(x\) approaches \(\pm\infty\). Importantly, like many established activation functions (e.g., ReLU, PReLU, GELU, SiLU), our approximation ensures that the activation function intersects the origin, introducing a third breakpoint at \(x = 0\). To capture GELU's capability for generating negative outputs, a breakpoint at \(x = -0.75\) approximates its minimum value, enhancing the fidelity of our approximation. To optimize the representation of GELU's transitional non-linear behavior within the intervals 
\([-3,-0.75]\) and \([0,3]\), additional breakpoints at \(x=-2.1\) and \(x=0.5\) are  introduced. These points were determined through the minimization of the mean square error, ensuring a more accurate approximation in the specified ranges. With the mentioned breakpoints, figure \ref{fig:gelu} visualizes our final approximation which is described in the equation below:
\[
\scriptsize
PEANO-GELU(x) = 
\left\{
	\begin{array}{ll}
		0 & \quad \text{if } x < -3 \\
		-0.0414(x+3)  & \quad \text{if } -3 \leq x < -2.1 \\
            -0.0982(x+2.1)-0.0373& \quad \text{if } -2.1 \leq x < -0.75 \\
             0.2266(x+0.75)-0.17 & \quad \text{if } -0.75 \leq x < 0 \\
             0.6914x  & \quad \text{if } 0 \leq x < 0.5\\
             1.0617(x-0.5) + 0.3457  & \quad \text{if }  0.5 \leq x < 3\\
             x  & \quad \text{if }   x \geq 3
	\end{array}
\right.
\]

\subsection{FPGA Implementation}
\label{sub:fpga-implement}
\begin{figure*}[htbp]
\centering
\includegraphics[width=0.8\textwidth]{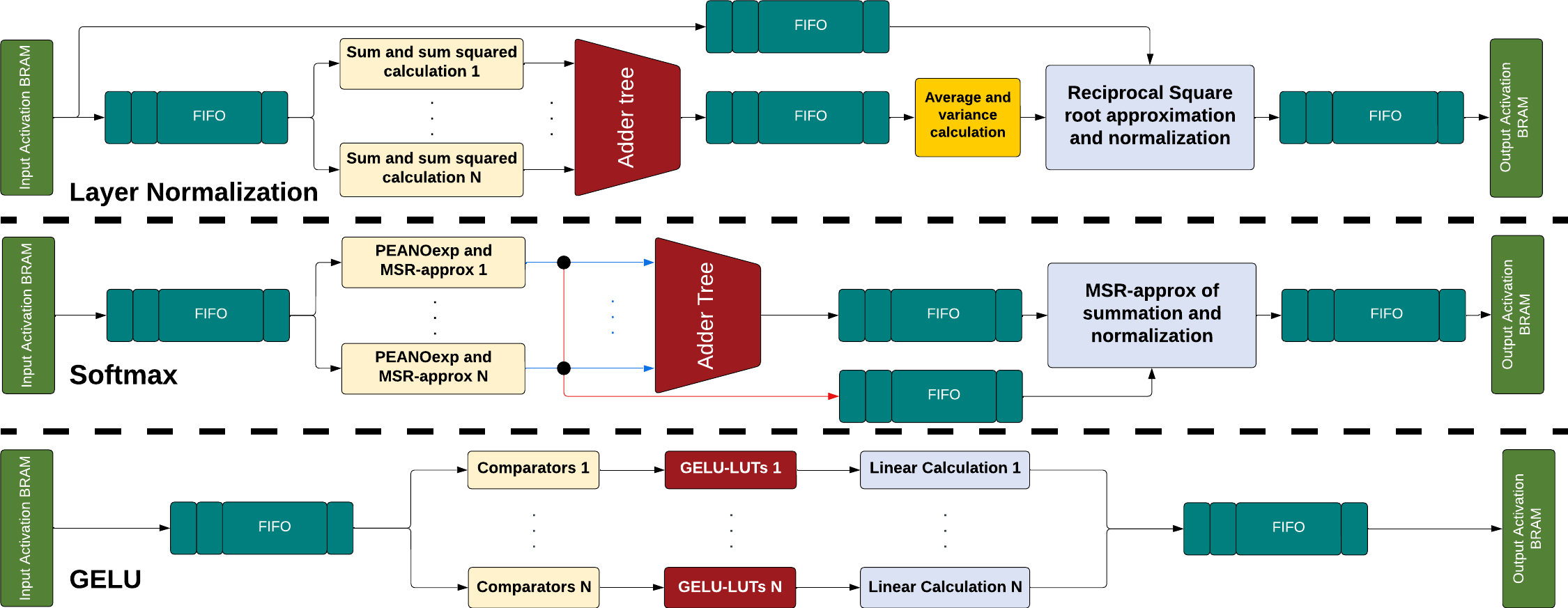}
\caption{Overall FPGA implementation of PEANO-ViT}
\label{fig:fpga-PEANO}
\vspace{-2mm}
\end{figure*}

The overall FPGA implementation of PEANO-ViT's non-linear layers is illustrated in Figure \ref{fig:fpga-PEANO}. Notably, each non-linear function processes \(N\) elements concurrently, enabling an approximate \(N-fold\) reduction in computation time. To enhance processing speed further, FIFO queues have been integrated between the reading, storing, and computing stages across all three implementations. Distinct from GELU, both layer normalization and softmax necessitate dual readings of input data—the initial for preliminary calculations and the subsequent for the normalization phase. Integrating an extra FIFO in parallel to the primary data stream notably decreases the latency for both the layer normalization and softmax modules by eliminating the requirement to temporarily store input values for a second calculation phase. Increasing the parameter \(N\) accelerates the processing of non-linear functions at the cost of more FPGA resource consumption. Consequently, PEANO-ViT becomes a configurable hardware framework alongside its software flexibilities.

\begin{algorithm}[tb]
    \caption{Multi-Scale Reciprocal approximation (MSR-approx)}
    \label{alg:base recip}
    \begin{algorithmic}[1]
   
        \Require $x, \alpha^*, StoredRecip[2^{\alpha^*+1}-1]=\{\frac{1}{1},\ldots,\frac{1}{2^{\alpha^*+1}-1}\}$
        \Ensure $y$
        \Comment{approximation of $\frac{1}{x}$ }
        \State $logInterval = LeadingOne(x)$
        \If{$logInterval \leq \alpha^*$}
            \State $\alpha = 0$
        \Else
            \State $\alpha = logInterval-\alpha^*$
        \EndIf
        \State $Scale = 2^{\alpha}$
        \State $y = ({StoredRecip[\lfloor x >> \alpha\rfloor]})>>\alpha$
        \State \Return $y$
    \end{algorithmic}
\end{algorithm}

\begin{algorithm}[bt]
   
    \caption{PEANO Softmax}
    \label{alg:PN-SFTMX}
    \begin{algorithmic}[1]
   
        \Require $x_1,\ldots, x_n$
        \Ensure $y_1,\ldots, y_n$
        \State $MaxInput = max(x_i)$
        \Comment{Maximum of inputs}
        \State $\tilde{x}_i = x_i - MaxInput + 2$
        \Comment{Shifting inputs by 2 - MaxInput}
        \For{$i=1$ {\bfseries to} $n$}
                \If{$\tilde{x}_i < -3$}
                    \State ${PEANOexp_i} = 0$
                \Else
                   \State ${PEANOexp_i} = (12+6\tilde{x}_i+\tilde{x}_i^{2})$\\
                    \quad\quad\quad\quad\quad\quad $\times$ MSR-approx$(12-6\tilde{x}_i+6\tilde{x}_i^{2})$
                   
                \EndIf
            \EndFor
        \State $Sum = \sum_{i=1}^{n} PEANOexp_i$ \Comment{Summation of exponential terms}
            \For{$i=1$ {\bfseries to} $n$}
                \State $y_i = PEANOexp_i \times $MSR-approx$(Sum) $
            \EndFor
        \State \Return $y_1,\ldots, y_n$
    \end{algorithmic}
\end{algorithm}

\begin{table}[tb]
\centering
\caption{Accuracy Loss of approximations on ImageNet-1K benchmark. The results of \cite{DBLP:conf/iccad/WangZSSL23} and \cite{li2023high}, if available, are directly sourced from the papers. FP32 and FiP16 stand for 32-bit floating-point and 16-bit fixed-point, respectively.}
\vspace{-2mm}
\label{tab:accuracy}
\resizebox{\columnwidth}{!}{
\begin{tabular}{c c c c}

\toprule
\textbf{Model}  & \textbf{Approach}    & \textbf{Approximations} & \textbf{Accuracy}  \\

\midrule[\heavyrulewidth]
\multirow{4}{*}{\textbf{DeiT-S}}  & Baseline(FP32)  & 
{-}  & {79.85\%}     \\
{}  & SOLE \cite{DBLP:conf/iccad/WangZSSL23}(FP32)  & 
{Layer normalization + softmax}    & {79.27\%} \\
{} & PEANO-ViT(Ours)(FP32) & {Layer normalization + softmax}& {\textbf{79.36\%}}  \\
{} & PEANO-ViT(Ours)(FiP16) & {All non-linearities} & {\textbf{79.13\%}} \\

\midrule[\heavyrulewidth]
\multirow{5}{*}{\textbf{DeiT-B}}  & Baseline(FP32)  & 
{-}    & {81.85\%}    \\
{}  & SOLE \cite{DBLP:conf/iccad/WangZSSL23}(FP32)  & 
{Layer normalization + softmax}   & {81.60\%}    \\
{} & PEANO-ViT(Ours)(FP32) & {Layer normalization + softmax}& {81.55\%}  \\
{} & PEANO-ViT(Ours)(FiP16) & {All non-linearities} & {81.35\%} \\
{} & PEANO-ViT(Ours) W LMSR-approx(FiP16) & {All non-linearities} & {\textbf{81.65\%}} \\

\midrule[\heavyrulewidth]
\multirow{4}{*}{\textbf{Swin-B}}  & Baseline(FP32)  & 
{-}   & {83.60\%} \\
{}  & SOLE \cite{DBLP:conf/iccad/WangZSSL23}(FP32)  & 
{Layer normalization + softmax}    & {83.05\%}\\
{} & PEANO-ViT(Ours)(FP32) & {Layer normalization + softmax}& {\textbf{83.60\%}}\\
{} & PEANO-ViT(Ours)(FiP16) & {All non-linearities}& {\textbf{83.56\%}}\\

\midrule[\heavyrulewidth]
\multirow{4}{*}{\textbf{ViT-L}}  & Baseline(FP32)  & 
{-}   & {85.15\%}\\
{}  &Li et al.\cite{li2023high}(FiP16)  & 
{Softmax + GELU}   & {84.78\%} \\
{} & PEANO-ViT(Ours)(FiP16) & {Softmax + GELU} & {\textbf{85.03\%}}\\
{} & PEANO-ViT(Ours)(FiP16) & {All non-linearities} & {\textbf{84.83\%}}\\

\bottomrule
\end{tabular}}
\vspace{-3mm}
\end{table}


\section{Results and Discussions}
\label{sec:res}
In this study, the PEANO-ViT model was implemented on a Xilinx UltraScale+ VU9P board running at a frequency of 250 MHz. We utilized the Vivado power report from Xilinx to evaluate the power consumption of each design. To evaluate the performance of PEANO-ViT, we employed the publicly available ImageNet-1K dataset \cite{DBLP:conf/cvpr/DengDSLL009} and three different model architectures, namely ViT \cite{DBLP:conf/iclr/DosovitskiyB0WZ21}, DeiT \cite{DBLP:conf/icml/TouvronCDMSJ21} and Swin \cite{DBLP:conf/iccv/LiuL00W0LG21}, across various sizes (small, base, and large). It is important to point out that our experimental setup does not require extensive retraining. Instead, we conducted only two epochs of fine-tuning after integrating each approximation into the model. We utilized pre-trained models from the TIMM library \cite{rw2019timm} as our starting point and implemented our approximations using PyTorch.
 

\begin{table*}[tb]
\centering
\caption{Hardware metrics for DeiT-B Implementation}
\vspace{-2mm}
\label{tab:hardware}
\resizebox{\textwidth}{!}{
\begin{tabular}{c c c c c c c c c}

\toprule
\textbf{Non-linear layer}   &\textbf{Approach} & \textbf{DSP} & \textbf{DSP (Reduction)} &  \textbf{LUT}& \textbf{LUT (Reduction)} & \textbf{Register}& \textbf{Register (Reduction)} &\textbf{Power efficiency} \\
\midrule[\heavyrulewidth]
\multirow{3}{*}{\textbf{Layer normalization}} & {Standard layer normalization} & {51} & {-}   &{24609}& {-}  &{29831}& {-}  &{$1\times$}  \\
{} & {LTrans-OPU \cite{DBLP:conf/fpl/BaiZZZCYW23}} & {0}& \textbf{100\%}    &{60902}& {-147.4\%}  &{7850}& \textbf{73.6\%}  &{$0.99\times$}  \\
{} & {PEANO layer normalization (Ours)} & {52}& {-1.9\%}    &{8157}& \textbf{66.8\%}  &{8621}& {71.1\%}  & {$\boldsymbol{1.91\times}$}  \\

\midrule[\heavyrulewidth]
\multirow{4}{*}{\textbf{Softmax}}  &{Standard softmax} & {64}& {-}   & {9745}& {-}  &{10648}& {-}  &{$1\times$}\\
{}  &{LTrans-OPU \cite{DBLP:conf/fpl/BaiZZZCYW23}} & {0}& \textbf{100\%}   & {238569}& {-2348.1\%}  &{13837}& {-29.9\%}  &{$0.19\times$}\\
{}  &{PEANO softmax W MSR-approx (Ours)} & {48}& {25\%}   & {5595}& \textbf{42.5\%}  &{3831}& \textbf{64\%}  &{$\boldsymbol{1.39\times}$}\\
{}  &{PEANO softmax W LMSR-approx (Ours)} & {49}& {23.4\%}   & {5741}& {41.1\%}  &{3876}& {63.6\%}  &{$1.38\times$}\\

\midrule[\heavyrulewidth]
\multirow{3}{*}{\textbf{GELU}} & {Standard GELU} & {128}& {-}   & {101267}& {-}  &{88293}& {-}  &{$1\times$} \\
{} & {LTrans-OPU \cite{DBLP:conf/fpl/BaiZZZCYW23}} & {0}& \textbf{100\%}    &{11314}& {88.8\%}  &{2499}& \textbf{97.1\%}  &{$6.76\times$} \\
{} & {PEANO GELU (Ours)} & {16}& {87.5\%}    &{2940}& \textbf{97.1\%}  &{2951}& {96.6\%}  & $\boldsymbol{8.01\times}$ \\
\bottomrule
\end{tabular}}
\end{table*}

\subsection{ImageNet Classification}
\label{subsec:class_acc}
Table \ref{tab:accuracy} provides a comparison of accuracy losses for four ViT-based models utilizing the PEANO-ViT approximations against techniques proposed by \cite{DBLP:conf/iccad/WangZSSL23} and \cite{li2023high} implemented on FPGA and GPU platforms, respectively. In our analysis, we set the layer normalization parameter \(m = 4\) and the MSR-approximation parameter \(\alpha^{*} = 4\) without any linear interpolation. The superior performance of PEANO-ViT compared to \cite{li2023high} and \cite{DBLP:conf/iccad/WangZSSL23} stems from its independent approximations of the softmax, GELU, and layer normalization functions, while \cite{li2023high} focuses solely on softmax and GELU, and \cite{DBLP:conf/iccad/WangZSSL23} on layer normalization and softmax. The results of Table \ref{tab:accuracy} indicate that PEANO-ViT exhibits minimal accuracy degradation when applying approximations to all non-linear blocks. Furthermore, when using a similar approximation approach, PEANO-ViT achieves lower accuracy reduction across DeiT-S, Swin-B, and ViT-L models compared to the methods outlined in \cite{li2023high} and SOLE \cite{DBLP:conf/iccad/WangZSSL23}. For the DeiT-B model, PEANO-ViT shows reduced accuracy degradation compared to SOLE \cite{DBLP:conf/iccad/WangZSSL23} when switching from MSR-approximation to LMSR-approximation. Notably, PEANO-ViT offers the ability to further minimize accuracy loss by adjusting \(m\) and \(\alpha^{*}\) and by incorporating linear interpolation in the MSR approximation (LMSR-approx).

\subsection{Hardware Cost}
\label{subsec:hard_cost}
Table \ref{tab:hardware} details the power efficiency gain and reduction in resource usage achieved by implementing PEANO-ViT. By utilizing the rapid and hardware-compatible approximations introduced by PEANO-ViT, the significant power consumption and resource usage associated with hardware-intensive and costly iterative methods for exact non-linear implementation have been greatly diminished. Furthermore, Table \ref{tab:hardware} provides the resource utilization breakdown for each non-linear layer of PEANO-ViT. In processing layers such as normalization, softmax, and GELU, we simultaneously handle 16 elements, resulting in a Level of Parallelism (LoP) of 16 to enable a fair comparison with LTrans-OPU. This LoP can be adjusted to align with resource availability and latency objectives, making PEANO-ViT a versatile framework for enhancing the speed of machine learning tasks. Increasing the LoP enhances processing speed but may lead to higher resource consumption and power usage.

\subsection{Flexibility of PEANO-ViT}
\label{subsec:flexibility}
PEANO-ViT is a highly versatile framework that can be tailored to meet specific accuracy goals, hardware resource limitations, and power consumption requirements. This adaptability is achieved through the adjustment of key parameters such as \(m\) for layer normalization, \(\alpha^{*}\) for softmax, and the selection between MSR or LMSR approximations for softmax. Furthermore, the framework offers flexibility in determining the number of linear segments for approximating the GELU function. Table \ref{tab:flex-MSE} illustrates the impact of different configurations on the mean square error accuracy of approximated functions. Increasing the values of \(m\) and \(\alpha^{*}\), expanding the number of linear segments in GELU, and choosing LMSR over MSR result in improved accuracy but also consume higher hardware resources, resulting in increased power consumption.

\begin{table}[tb]
\centering
\caption{Effect of PEANO-ViT parameters on approximations accuracy}
\vspace{-4mm}
\label{tab:flex-MSE}
\resizebox{\columnwidth}{!}{
\begin{tabular}{c c c c}

\toprule
\textbf{Fuction}  & \textbf{Test input interval}    & \textbf{Changed parameter} & \textbf{MSE}  \\

\midrule[\heavyrulewidth]
\multirow{3}{*}{\textbf{Reciprocal square root}}  & \multirow{3}{*}{$[1,128]$}  & 

{$m = 3$}  & {$4.93\times10^{-5}$}     \\
{}  & {}  & 
{$m = 4$}    & {$9.56\times10^{-6}$} \\
{}  & {}  & 
{$m = 5$}    & {$7.86\times10^{-6}$} \\

\midrule[\heavyrulewidth]
\multirow{4}{*}{\textbf{Reciprocal}}  & \multirow{4}{*}{$[8,64]$}  & 
{$\alpha^{*} = 4, MSR$}    & {$4.19\times10^{-6}$}    \\
{}  & {}  & 
{$\alpha^{*} = 5, MSR$}   & {$4.03\times10^{-6}$}    \\
{} & {} & {$\alpha^{*} = 4, LMSR$}& {$3.63\times10^{-9}$}  \\
{} & {} & {$\alpha^{*} = 5, LMSR$} & {$3.58\times10^{-9}$} \\

\midrule[\heavyrulewidth]
\multirow{2}{*}{\textbf{GELU}}  & \multirow{2}{*}{$[-4,4]$}  & 
{7 segments}   & {$2.65\times10^{-4}$} \\
{}  & {}  & 
{10 segments}    & {$8.31\times10^{-5}$}\\

\bottomrule
\end{tabular}}
\vspace{-3mm}
\end{table}

\section{Conclusion} 
\label{sec:conc}
PEANO-ViT optimizes ViT models by approximating non-linear blocks and eliminating division operations, maintaining high accuracy with minimal reduction. This approach enhances power efficiency and resource savings, setting a new benchmark for sustainable deep learning. Its flexibility allows for customized adjustments in accuracy, hardware resources, and power consumption, ensuring it meets specific performance requirements without sacrificing efficiency or accuracy.

\textbf{Acknowledgment:} This research is supported by a grant from the Software and Hardware Foundations program of the NSF.

\bibliographystyle{ACM-Reference-Format}
\bibliography{PEANO}

\end{document}